\pdfoutput=1
\documentclass[11pt]{article}

\usepackage[]{acl}

\usepackage{times}
\usepackage{latexsym}

\usepackage[T1]{fontenc}
\usepackage[utf8]{inputenc}
\usepackage{enumitem}
\usepackage{microtype}

\usepackage{inconsolata}

\usepackage{times}
\usepackage{latexsym}
\usepackage{bm}
\usepackage[T1]{fontenc}
\usepackage[utf8]{inputenc}
\usepackage{microtype}
\usepackage{inconsolata}
\usepackage{multirow}
\usepackage{graphicx}
\usepackage{booktabs}  
\usepackage{subcaption} 
\usepackage{xspace}
\usepackage{xcolor}
\usepackage{hyperref}
\usepackage{comment}

\definecolor{green}{HTML}{38761D}
\definecolor{orange}{HTML}{FF8900}
\definecolor{blue}{HTML}{000CC7}

\newcommand{\lens}{\texttt{LegalLens}\xspace} 
\newcommand{\ner}{\texttt{LegalLens-NER}\xspace}
\newcommand{\nli}{\texttt{LegalLens-NLI}\xspace}

\title{LegalLens Shared Task 2024: Legal Violation Identification in Unstructured Text}

\author{%
\textbf{Ben Hagag}$^{\textcolor{blue}{\star}}$
Liav Harpaz$^{\textcolor{blue}{\star}}$\hspace{3mm}%
\textbf{Gil Semo}$^{\textcolor{blue}{\star}}$\hspace{3mm}%
\textbf{Dor Bernsohn}$^{\textcolor{blue}{\star}}$\hspace{3mm}%
\textbf{Rohit Saha}$^{\textcolor{green}{\dag}}$\hspace{3mm}%
 \\
\textbf{Pashootan Vaezipoor}$^{\textcolor{green}{\dag}}$\hspace{3mm}%
\textbf{Kyryl Truskovskyi}$^{\textcolor{red}{\ddagger}}$\hspace{3mm}%
\textbf{Gerasimos Spanakis}$^{\textcolor{purple}{\phi}}$ \\
\\
$^{\textcolor{blue}{\star}}$Darrow AI Ltd., Tel Aviv, Israel \texttt{\{firstname.lastname\}@darrow.ai} \\
$^{\textcolor{green}{\dag}}$Georgian.io, Toronto, Canada \texttt{\{firstname\}@georgian.io} \\
$^{\textcolor{red}{\ddagger}}$Scoreinforce, Toronto, Canada \texttt{\{firstname\}@Scoreinforce.com}\\
 $^{\textcolor{purple}{\phi}}$Maastricht University, Maastricht, Netherlands \texttt{jerry.spanakis@maastrichtuniversity.nl}
}
\begin{document}
\maketitle
\begin{abstract}
%In this paper we present \lens shared task consisting two main subtasks: first is \ner for detecting legal violations within unstructured textual data, and the second is \lens-NLI for associating these violations with potentially affected individuals. The two subtasks based on \lens novel datasets. For Subtask A the performances of top-ranked systems are close to that of humans. However,
%for Subtask B, there is still a relatively large gap between systems and human performance. 
%The dataset used in our task can be found at https://huggingface.co/datasets/darrow-ai/\lensNLI-SharedTask 
%The leaderboard can be found at
%https://www.codabench.org/competitions/3052.

This paper presents the results of the LegalLens Shared Task, focusing on detecting legal violations within text in the wild across two sub-tasks: \ner for identifying legal violation entities and \nli for associating these violations with relevant legal contexts and affected individuals. Using an enhanced LegalLens dataset covering labor, privacy, and consumer protection domains, 38 teams participated in the task. Our analysis reveals that while a mix of approaches was used, the top-performing teams in both tasks consistently relied on fine-tuning pre-trained language models, outperforming legal-specific models and few-shot methods. The top-performing team achieved a 7.11\% improvement in NER over the baseline, while NLI saw a more marginal improvement of 5.7\%. Despite these gains, the complexity of legal texts leaves room for further advancements.

% Our analysis reveals key trends in the approaches used: while participants employed a combination of few-shot learning and fine-tuning methods, the top-performing groups in both tasks relied on fine-tuning. Interestingly, models fine-tuned on pre-trained language models specifically tailored for legal texts outperformed those fine-tuned on general language models, highlighting the importance of domain-specific adaptation in achieving superior results for legal violation detection.

\end{abstract}

\section{Introduction}

Legal violations are everywhere, but often go unnoticed. In many areas such as privacy, consumer protection, environmental law, and labor regulations, traces of these violations, indicating wrongdoing, are frequently lost in the vast amounts of digital information. As the world becomes increasingly digital, it is inevitable that traces of these legal violations can be found online. This concept is the foundation of the \lens project \cite{bernsohn2024legallens}. These violations pose significant risks to individuals and institutions, undermining legal and ethical standards in our increasingly digital society. Therefore, developing advanced methods to detect and address these violations is crucial.

Identifying legal violations on the open web presents two primary challenges: first, determining where to search, and second, accurately interpreting whether the information indicates a legal violation. The first challenge involves going through massive amounts of online content, selecting sources that are likely to yield relevant information while accounting for varying levels of credibility and relevance. The second challenge lies in applying legal knowledge and to determine the legal grounds for these potential violations, and identify victims who may be entitled to compensation.

\begin{figure}[ht]
    \centering
    \includegraphics[width=0.45\textwidth]{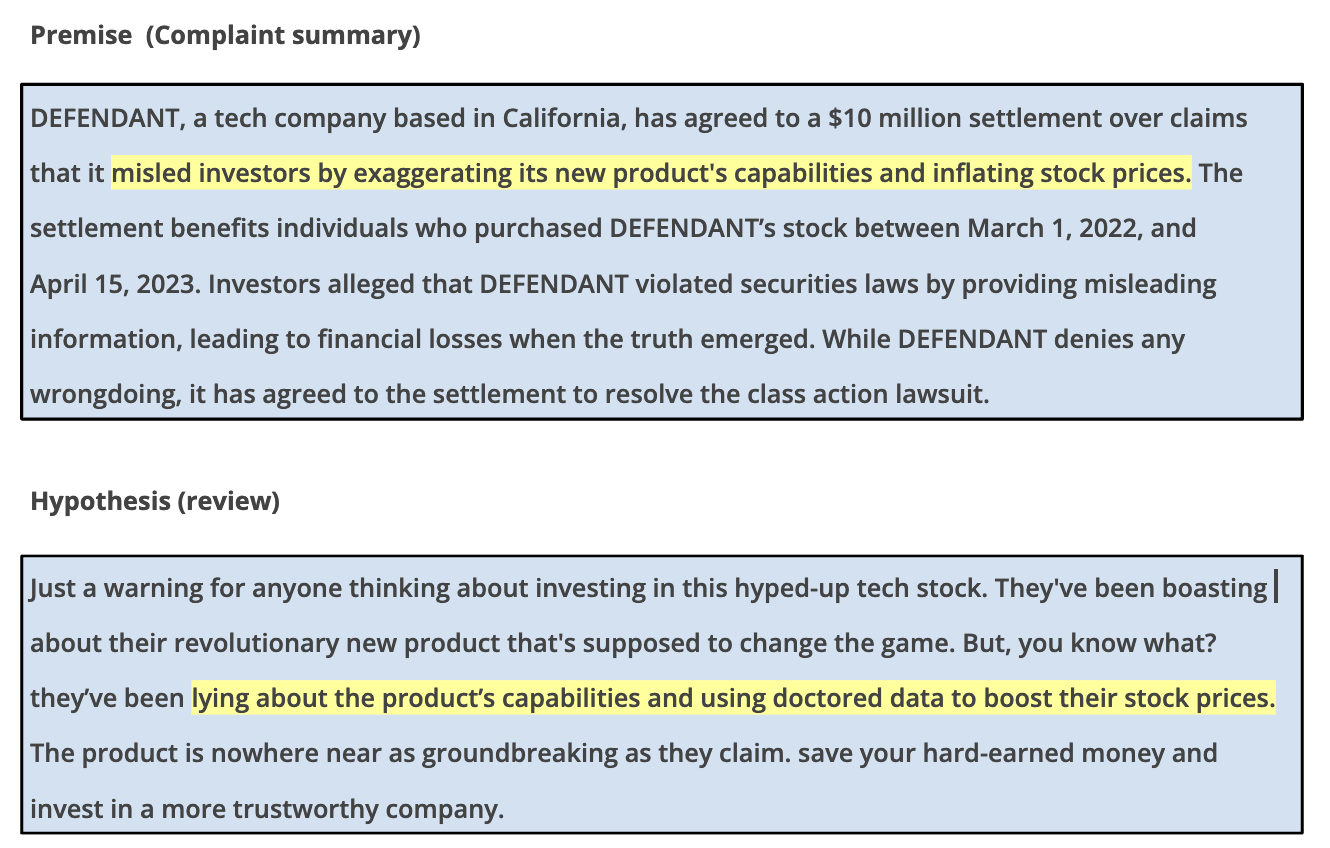}
    \captionsetup{width=0.9\linewidth} 
    \caption{Example of the NLI sub-task showing how premises like court-filed complaints or articles are used to identify individuals harmed by violations. Both the premise and hypothesis were selected due to matching violation entities identified by the LegalLens-NER model, illustrating the system's ability to link legal grounds to personal experiences and recognize potential victims.}
    \label{fig:NER_example}
\end{figure}

To advance this field, and to address these challenges, \lens tasks were presented in \cite{bernsohn2024legallens}. The underlying assumption of \lens is that Legal violations often leave digital traces, which can be uncovered through careful analysis. \lens presented a two-step approach to tackle these challenges: The first is \ner (Named Entity Recognition) to extract legal violation entities from online data. The task involves detection and categorization of specific legal violation entities such as laws, violations, violators, and victims within unstructured text. Simple NER methods do not focus on these types of entities and fail to capture the ambiguity of legal language. Figure \ref{fig:NER_example} shows an example of the NER task.

\begin{figure}[htbp]
    \centering
    \includegraphics[width=0.45\textwidth]{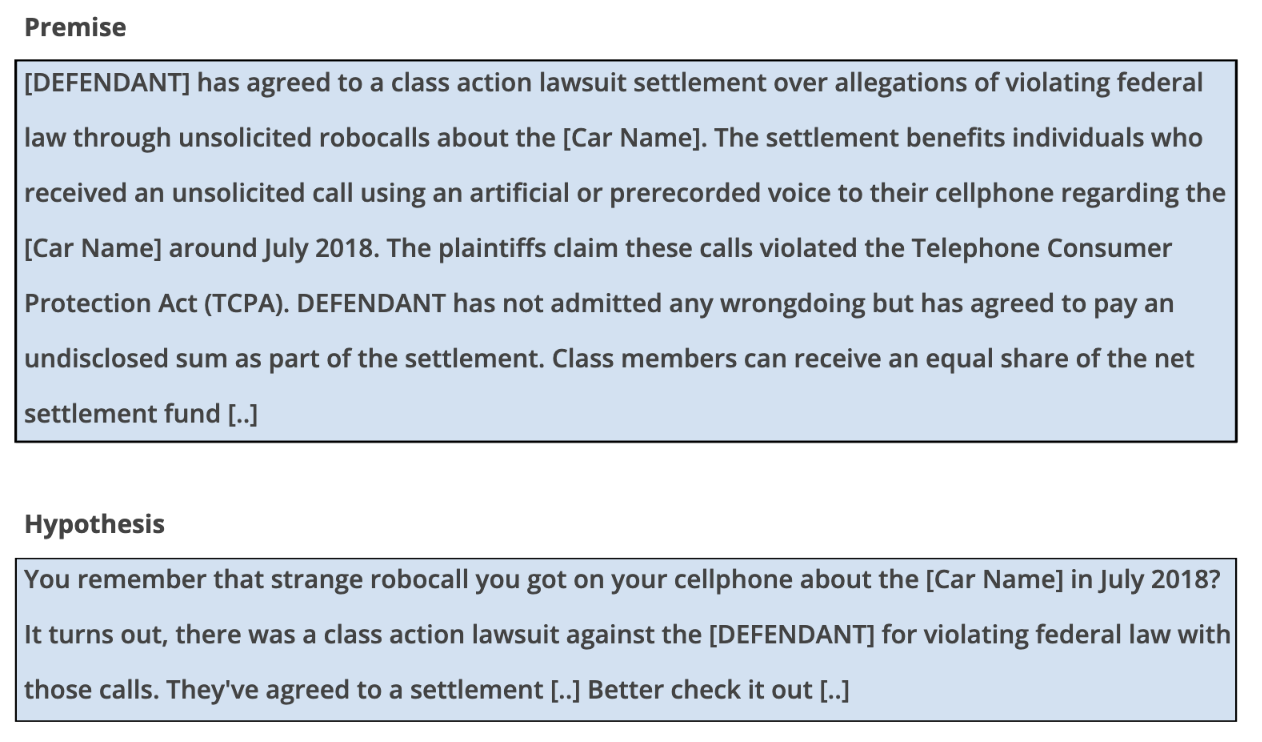}
    \captionsetup{width=0.9\linewidth} 
    \caption{An example of the LegalLens NLI task, where the model assesses whether the provided hypothesis (a potential legal violation) is supported, contradicted, or unrelated to the premise (an allegation summary).
}
    \label{fig:nli_entailed example}
\end{figure}

The second step is the \nli (Natural Language Inference) to associate identified violations with relevant legal cases or statutes. More specifically, the task given a premise (allegation summary of a legal case) determine the relationship to a hypothesis (a potential detected violation) and classify their relationship as entailment, contradiction, or neutrality. Figure \ref{fig:nli_entailed example} shows an example of the NLI task.

The datasets for these two sub-tasks were built upon proprietary data by Darrow.ai\footnote{https://www.darrow.ai/}, designed to be as realistic as possible and to capture the nuances and variability of real-world cases. The data was generated in utilizing GPT-4o \cite{openai2023gpt4} and domain experts, ensuring both realism and complexity.

The 1st Shared Task on \lens was organized to encourage new research at the intersection of natural language processing and legal studies and to stimulate interest in legal violation detection within the NLP community.

In this paper, we present the results of the Shared Task, offering a detailed description of the evaluation data and the systems developed by participants. We analyze the performance of the participating systems, evaluating their capabilities in processing legal language and identifying legal violations. The top-performing systems for NER showed a substantial improvement over the baseline, with a 7.11\% increase in F1 score for the best team. The NLI task saw more marginal progress, with only one team outperforming the baseline by 5.7\%. While these improvements highlight progress in legal violation detection, particularly in entity recognition, there remains significant room for further advancements in handling the complexities of natural legal language inference.

As a result, this shared task holds value not just for experts in Machine Learning and NLP, but also for legal professionals, sociologists, and policymakers. This initiative has the potential to foster interdisciplinary collaborations and contribute to advancements in detecting legal violations in the digital era. We are happy to see interdisciplinary teams with participants from CS and NLP alongside legal practitioners, students and social science. 

The remainder of this paper is structured as follows: In Section 2, we provide an overview of the \lens tasks. Section 3 describes our data collection process, while Section 4 presents the systems and results. Section 5 delves into the details of the three winning teams, and Section 6 offers an overview of the current research landscape.

\begin{figure}[ht]
    \centering
    \includegraphics[width=0.5\textwidth]{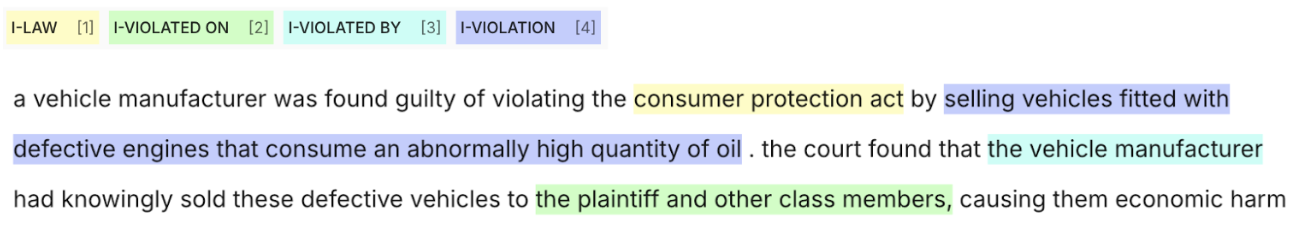}
    \caption{NER sub-task example showing highlighted legal violation entities, including Law, Violation, Violation By, and Violation On.}
    \label{fig:ner explicit example}
\end{figure}

\section{What is \lens}

To efficiently detect legal violations across various domains in the online digital data, a system must be developed that can scan large datasets, isolate relevant information, and accurately map it to appropriate legal grounds. This involves scanning large amount of online data, contextualizing the findings by linking them to specific legal grounds, and clearly explaining potential violations. Additionally, the system must identify the affected individuals or entities who may be entitled to legal recourse, thereby enabling effective enforcement and remediation.

\lens is designed to address these challenges by providing a structured approach to detecting legal violations in digital data. It achieves this by breaking down the task into two key components: \ner to identify relevant legal entities and \nli to determine the relationship between data points and legal grounds. In the following section, we will delve deeper into each of these sub-tasks, explaining how they contribute to the overall goal of efficiently detecting and contextualizing legal violations. For the full description of both tasks please refer to the original \lens work \cite{bernsohn2024legallens}.

\subsection{\ner}
The \ner task in the \lens framework aims to identify key legal entities relevant to detecting violations in unstructured text. The \ner model classifies tokens into predefined categories: \textit{Law} (the specific law violated), \textit{Violation} (the nature of the infraction), \textit{Violated By} (the responsible entity), and \textit{Violated On} (the affected party).

Given the sheer volume of data and the challenge of identifying relevant information, the \ner task acts as an initial filter, extracting critical entities like laws and violations while discarding irrelevant or non-essential information. This process ensures that only the most pertinent data is selected for further analysis and association with legal grounds in subsequent tasks (\nli).  The primary goal is to highlight relevant data points for deeper legal examination, making the subsequent steps more efficient and focused.

The dataset for the \ner task was curated from class action complaints, with the key violation sections extracted and then summarized and refined using GPT-4. The generated text was formatted as articles, reviews, and social media posts. Human experts validated the realism of the output and annotated the entities. Two prompting strategies were used: explicit, focusing on multiple entities with specific structure, and implicit, centering on a single entity, particularly the violation content. Additional parameters like cause of action and industry were also included to tailor the content to various and real-world scenarios.

\subsection{\nli}

The NLI task in the \lens framework is designed to map identified legal violations to the most relevant Cause of Action (CoA) or legal statute. By proceeding the \ner model, this model aligns the relevant data with specific legal frameworks, such as laws or precedents, and provides a justification or reasoning for the connection between the identified violations and the applicable legal grounds.

As we worked on this task, we understood that \nli can serve another important purpose: identifying individuals who may have been harmed by the violation. By using descriptions of violations—such as court-filed complaints, articles, or other texts—as premises, the NLI task can analyze relevant online content, like reviews or posts, where people describe their personal experiences. This allows the system to link the identified legal violations to specific individuals who have suffered harm, thus expanding its capability to both identify legal grounds and recognize victims of these violations.

This combined approach strengthens the process of tracing violations back to real-world consequences, making it possible to identify affected individuals with greater precision and relevance.

The dataset is derived from curated legal news articles, with the key violation sections summarized and refined using LLM. The premise for each sample consists of these curated and summarized violation descriptions. The hypotheses were generated using different LLM setups to simulate various scenarios and complaints to reflect real-world situations.

The dataset is labeled with three categories: \textit{Entailment} where the violation is directly supported by the legal grounds; \textit{Contradiction}, where the violation contradicts the legal grounds; and \textit{Neutral}, where the relationship between the violation and the legal grounds is ambiguous or irrelevant.

Human experts validated the correctness and completeness of the premises and hypotheses, and annotated the NLI labels accordingly.

% The dataset is labeled with three categories: Entailment, where the violation is directly supported by the legal grounds. Contradiction, where the violation contradicts the legal grounds, and Neutral, where the relationship between the violation and the legal grounds is ambiguous or irrelevant.

\section{Dataset Curation for \lens Shared Task}

Building upon the original \lens dataset and addressing its limitations, we have created a more comprehensive and challenging benchmark for \lens sub-tasks.

 Our goal was to create a dataset that not only mimics real-world scenarios but also presents a challenging benchmark for state-of-the-art NLP models in the legal domain.

The resulting dataset for the shared task maintains the dual-task structure of the original \lens, focusing on NER for violation identification and NLI for matching violations with known cases.  
With improved prompt practices, better annotators guidelines, human expert practices, and following feedback from the original paper\cite{bernsohn2024legallens}, we improve the generation process and the resulting annotations, yielding more realistic content, improve data quality and reduce bias.

Our enhancement process consisted of three primary steps:
The first was to clean the dataset from duplicated or almost duplicates examples. In some cases we have found that similar patterns appears in the dataset too often that is not making sense. We have tried to detect these patterns and exclude instances that repeat them. Also, to prevent potential biases and ensure broader applicability, we masked all company names within the dataset, including Defendants and Plaintiff names in the NLI dataset. We found that models were prone to overfitting if this masking process was not applied.

Additionally, we have implemented an improved three-step validation where legal experts conducted a multi-stage validation process, including a review for factual accuracy and legal relevance, cross-validation of NER annotations, and examination of premise-hypothesis pairs for logical consistency, completeness and correctness in the NLI task. 
All annotation conducted via Argilla \cite{Daniel_Argilla_-_Open-source_2023} available under an Apache-2.0 license.

Table \ref{table:ner_entities} shows the datasets tokens distribution. Also, table \ref{table:NLI_DOMAINS} shows the distribution of labeled samples across various legal domains for the NLI task, formatted as Contradiction/Entailment/Neutral. 
\label{sec:data_destribution}

\begin{table}[h]
\centering
\resizebox{0.5\textwidth}{!}{%
\begin{tabular}{|c|p{3.5cm}|c|}
\hline
\textbf{Entity} & \textbf{Description} & \textbf{\# Labeled Samples} \\
\hline
LAW & Specific law or regulation breached. & 373 \\
\hline
VIOLATION & Content describing the violation. & 1665 \\
\hline
VIOLATED BY & Entity committing the violation. & 373 \\
\hline
VIOLATED ON & Victim or affected party. & 373 \\
\hline
\end{tabular}
}
\caption{Distribution of NER entities generated through the combined dataset from the original paper and the updated process, excluding duplicates.}
\label{table:ner_entities}
\end{table}

\begin{table}[h]
\centering
\resizebox{0.5\textwidth}{!}{%
\begin{tabular}{|c|p{3.5cm}|c|c|}
\hline
\textbf{Entity} & 
\textbf{Description} & 
\textbf{Labels} & 
\textbf{\# Labeled Samples} \\
\hline
Consumer Protection & Deceptive advertising, fraud and unfair business practices. & 28/47/32 & 107 \\
\hline
Privacy & Unauthorized collection, use, or disclosure of personal data. & 80/72/82 & 234 \\
\hline
TCPA & Unauthorized telemarketing calls, faxes and text messages. & 38/34/39 & 111 \\
\hline
Wage & Illegal underpayment and unfair compensation practices by employers. & 9/7/5 & 21 \\
\hline
\end{tabular}
}
\caption{Distribution of labeled samples across various legal domains for the NLI task, formatted as Contradiction/Entailment/Neutral. This dataset combines samples from the original paper and the updated process, excluding duplicates.}
\label{table:NLI_DOMAINS}
\end{table}

\section{System Descriptions and Performance}
The competition was hosted on CodaBench\footnote{LegalLens shared task website: https://www.codabench.org/competitions/3052/} \cite{codabench}. During the evaluation phase, the leaderboard was hidden, meaning participants did not receive feedback on their submission scores until the phase concluded. Each team was allowed one submission per sub-task.

Both sub-tasks were evaluated as in the original paper: the \ner sub-task was assessed using the weighted F1 score, 
to account for class imbalance, with each class's F1 score weighted by the number of true instances. Evaluation was conducted using the seqeval\cite{seqeval} method, which requires exact matches between predicted and true entity spans—both the boundaries and the entity type must match precisely. We followed the IBO format (Inside, Beginning, Outside), where a correct match requires both the boundaries and tags to be accurate.
The \nli sub-task used the standard macro F1 score. Participants received the hidden test set only two days before the submission deadline, after submitting the source code of their best architecture. Changes to the model were not permitted after the release of the hidden test set.
During the evaluation phase, organizers verified that the predictions could be reproduced using the submitted source code.

\subsection{Baseline Systems}
As a baseline for the each sub-task, we use the best models from the original LegalLens paper \cite{bernsohn2024legallens}. We trained and evaluated the best models on the new datasets generated for the shared task, as described above. That is to make sure our baseline is up-to-date and performance improvement by participants is by better models, not just by our new dataset. For \ner the best model is RoBERTa-base which was fine tuned on the \ner dataset. The macro F1-Score for this model is 38.1\%. For \nli: the best model is Falcon-7B \cite{falcon40b} which achieved the highest score of 80.7\% macro F1 on average across domains.

\subsection{Participating teams}
A total of 87 individual users grouped in 38 teams
participated in the shared task, out of which the highest seven teams elected to write a system description paper. Most of the teams participated in both sub-tasks. Table \ref{table:NER_f1_scores} presents the results for the top six teams in the \ner sub-task, Table \ref{table:NLI_f1_scores} shows the results for the \nli sub-task, and Table \ref{table:NER_f1_scores_detailed} shows an entity level performance for the \ner sub-task. Most teams achieved better results than our baseline.
Another point worth noting is that success in one sub-task does not necessarily translate to success in the other. Out of the 38 teams, only the NowJ team made it to the top three systems in both tasks. This highlights that the challenges posed by the \ner and \nli sub-tasks are distinct, requiring different approaches and strengths.

Lastly, we note that there is a ceiling in terms of performances in the NER task. The top 4 teams achieve score around 70\% F1 score, which seems to be the plateau. suggesting that there is  room for improvement.

We present the leaderboard for both NER and NLI tasks, showcasing the top six teams and their F1 scores. The next section delves into the leading approaches in each task.

\begin{table}[h!]
    \centering
    \begin{tabular}{|l|c|}
        \hline
        \textbf{Team Name} & \textbf{Test F1 Score} \\
        \hline
        Nowj  & 0.416 \\
        \hline
        Flawless Lawgic & 0.402 \\
        \hline
        UOttawa & 0.402 \\
        \hline
        Baseline & 0.381 \\
        \hline
        Masala-chai & 0.380 \\
        \hline
        UMLaw\&TechLab & 0.321 \\
        \hline
        Bonafide & 0.305 \\
        \hline
    \end{tabular}
    \caption{Top six teams for the \ner sub-task,with performance measured by weighted F1 scores on a hidden test set.}
    \label{table:NER_f1_scores}
\end{table}
\begin{table}[h!]
    \centering
    \begin{tabular}{|l|c|}
        \hline
        \textbf{Team Name} & \textbf{Test F1 Score} \\
        \hline
        1-800-Shared-Tasks & 0.853 \\
        \hline
        Baseline & 0.807 \\
        \hline
        Semantists & 0.785 \\
        \hline
        Nowj & 0.746 \\
        \hline
        UOttawa & 0.724 \\
        \hline
        bonafide & 0.653 \\
        \hline
        masala-chai & 0.525 \\
        \hline
    \end{tabular}
    \captionsetup{width=0.9\linewidth} 
    \caption{Top six teams for the \nli sub-task,with performance measured by Macro F1 scores on a hidden test set.}
    \label{table:NLI_f1_scores}
\end{table}

In the NLI task, the leading team employed a Mixture-of-Experts approach\cite{jiang2024mixtral}, which significantly outperformed the subsequent teams. 

%It would be useful to analyze whether top predictions align with the most relevant expert model and assess how well this approach generalizes across legal domains.

All submitted models are available in Darrow.ai's Hugging Face Space\footnote{https://huggingface.co/darrow-ai}.

\begin{table}[h!]
\scriptsize
    \centering
    \begin{tabular}{|l|c|c|c|c|c|}
        \hline
        \textbf{Team} & \textbf{Law} & \textbf{Violation} & \textbf{V-By} & \textbf{V-On}  \\
        \hline
        Nowj & 0.7310 & 0.630 & 0.041 & 0.337 \\
        \hline
        Flawless Lawgic & 0.711  & 0.582  & 0.081  & 0.310 \\
        \hline
        UOttawa & 0.701  & 0.626  & 0.045  & 0.299 \\
        \hline
        Baseline  & 0.668 & 0.499 & 0.087 & 0.353\\
        \hline
        Masala-chai  & 0.636  & 0.589  & 0.042  & 0.308 \\
        \hline
        UMLaw\&TechLab   & 0.596  & 0.573  & 0.047  & 0.104 \\
        \hline
        
        Bonafide & 0.750  & 0.230  & 0.152  & 0.264 \\
        \hline
    \end{tabular}
    \caption{Entity-specific performance for each team in the \ner sub-task, showing F1 scores for the identification of Law, Violation, Violated-By, and Violated-On entities.}
    \label{table:NER_f1_scores_detailed}
\end{table}

\section{Deeper Analysis}
In this section, we describe the key methodologies and innovative techniques employed by the top-performing teams in the LegalLens Shared Task.

\subsection{\ner Methodologies Overview}

The NowJ team, which achieved the highest score in the \ner sub-task, with 0.416 weighted F1 score, adopted a methodical approach that involved data utilization, preprocessing, and model fine-tuning. They have leveraged both LegalLens-NER datasets, the one from the original paper, and the one introduced for the shared task. The former consisted of 710 training samples and 617 test samples, totaling 1,327 samples. The latter contained 976 samples. To optimize training, the team selected the 976 samples from the \texttt{LegalLensNER-SharedTask} as the training set, with the remaining 351 samples (that are not included in the original dataset) from the \texttt{LegalLensNER} dataset used as the validation set. The model architecture combined a pre-trained language model with a Conditional Random Field (CRF) layer. Pre-trained Language Model - the team used the \texttt{Legal Longformer} (\texttt{lexlms/legal-longformer-base})\cite{chalkidis-garneau-etal-2023-lexlms}, a transformer-based model specialized for legal text. This model produced contextualized word embeddings, which was used for capturing the semantic nuances of the input text. Conditional Random Field (CRF) Layer modeled dependencies between labels, to ensure valid label sequences by optimizing the Maximum Likelihood Estimate (MLE). The team implemented the forward\cite{blunsom2004hidden} and Viterbi\cite{forney1973viterbi} algorithms during training and inference to calculate the probabilities of label sequences and decode the most likely sequence, respectively. Training setting includes: LM: Legal lexlms/legal-longformer-base\cite{chalkidis-garneau-etal-2023-lexlms}, Max Sequence Length: 256, Initial Learning Rate: 5e-5, Learning Rate for CRF and Fully Connected Layer: 8e-5, Weight Decay (Fine-Tuning): 1e-5, Weight Decay (CRF and Fully Connected Layer): 5e-6, Batch Size: 16, Total Training Epochs: 30 (Best epoch: 18th), Warmup Proportion: 0.1.

To address the issue of subword tokens in the datasets, where subwords were predicted with the 'X' label, the team implemented a post-processing step. This involved replacing any 'X' label with the label of the preceding token. If the preceding token was a 'B-' (beginning) label, the 'X' label was converted to the corresponding 'I-' (inside) label, ensuring the sequence followed the correct labeling structure.

The uOttawa team, which achieved the third-best score in the \ner sub-task, with a 0.402 weighted F1 score, developed their model using the SpaCy library\cite{spacy2}. The team implemented preprocessing steps to clean and remove null values and to ensure each token had a corresponding NER tag. The team treated the tokens as features, a transformer model, \texttt{microsoft/deberta-v3-base} for contextual embedding, and a custom NER component via \texttt{Tok2Vec\cite{Honnibal_spaCy_Industrial-strength_Natural_2020}} layer, to represent tokens in a high-dimensional vector space to capture semantic similarities between words. The model's performance was evaluated after each epoch on a validation set to monitor over-fitting. 
\subsection{\nli Methodologies Overview}

The Bonafide team, which achieved the fifth highest score in the \nli subtask, developed a methodology involving data augmentation and model fine-tuning. They used Mixtral 8x7b-instruct-v0.1-hf model \cite{jiang2024mixtral} to generate paraphrases for both premises and hypotheses across the original 312 rows of data. The model was prompted to produce realistic rephrasings that retained all the details of the original text, resulting in a final dataset of 936 rows. For model training, the Bonafide team utilized the \texttt{sileod/deberta-v3-small-tasksource-nli} \cite{sileo2023tasksource} encoder, which is based on the DeBERTa-v3-small architecture. This encoder, fine-tuned on tasksource for 250,000 steps and oversampled for long NLI tasks, was further fine-tuned on the augmented dataset. The training dataset was tailored to each legal domain, comprising only synthetic data relevant to that domain, while the test dataset remained unaltered. The hyperparameters used for training included a batch size of 8, a learning rate of 2e-5, and a linear learning rate scheduler. The models were trained for 10 epochs with early stopping to optimize performance. Final predictions on the test dataset were derived by aggregating outcomes from four domain-specific models. The most confident label was selected by calculating the argmax on the confidence levels of all four models.

The 1-800-Shared-Tasks team, which achieved the highest score in the \nli sub-task, with 0.853 macro f1 score, implemented a method involving the use of the \texttt{FastLanguageModel} from the Unsloth library\footnote{https://github.com/unslothai/unsloth}. Their approach focused on fine-tuning the \texttt{PHI3-Medium-NLI-16bit} model, with specific configurations to optimize performance on the NLI task. The model was loaded with a maximum sequence length of 2048 and configured to operate in 4-bit precision to manage computational efficiency. They further enhanced the model using LoRA (Low-Rank Adaptation) adapters \cite{hu2021lora}, allowing for the fine-tuning of only 1\% to 10\% of the model's parameters.

The \textbf{NowJ} team, which achieved the third-best score in the \nli sub-task, utilized two datasets provided by the competition organizers on HuggingFace: \texttt{darrow-ai/LegalLensNLI} and \texttt{darrow-ai/LegalLensNLI-SharedTask}. Both datasets contained only a training split with 312 samples. Upon preprocessing, which included converting text to lowercase, removing punctuation, and eliminating extra spaces, they identified approximately 160 differing samples between the two datasets. To maximize data utilization, the participants created a unified dataset comprising the original 312, and the new 160 samples. The combined dataset was then split into training and validation sets, with a test size of 0.4, resulting in 283 examples for training (\texttt{train\_raw}) and 189 examples for validation. Additionally, augmented versions of the examples from the first dataset were appended to create an expanded training set: 665 examples for training set and 189 examples for validation set. The data augmentation implemented using LangChain \cite{Chase_LangChain_2022} and the GPT-4o-mini \cite{achiam2023gpt} model via API. The goal was to paraphrase both the hypotheses and premises to simulate varying levels of English language proficiency, specifically targeting IELTS\footnote{https://ielts.org/} levels 6.5 and 8.5\footnote{https://ielts.org/take-a-test/preparation-resources/understanding-your-score}. The dataset was expanded with columns to track original and augmented examples, distinguishing versions by IELTS levels. A Pydantic model ensured data consistency, while the GPT-4o-mini model was guided by structured prompts to generate paraphrases. A custom Paraphraser class managed the process, maintaining the integrity of the original meaning. The NowJ team conducted a thorough evaluation of state-of-the-art pre-trained models, including LegalBERT \cite{schalkidis-etal-2020-legal}, T5\cite{raffel2020exploring}, and DeBERTa, to identify the optimal architecture for the NLI subtask. DeBERTa (\texttt{MoritzLaurer/DeBERTa-v3-large-mnli-fever\newline-anli-ling-wanli}) emerged as the best-performing model due to its stability and high F1-macro scores across multiple training iterations.

Performance across teams varied significantly by legal domain. The 1-800-Shared-Tasks team for instance, performed exceptionally well in structured domains like Privacy (1.0 F1) and TCPA, yet underperformed in complex domains like Wage, likely due to its smaller dataset size and implicit nature of violations. Similarly, NowJ and UOttawa also struggled in domains like Wage, but Semantists fared better due to a more balanced approach across legal domains, highlighting differences in generalization capability across teams. Models fine-tuned on larger datasets showed better overall performance, but those specializing in domain-specific tasks demonstrated marginal improvements, revealing a gap in domain adaptation.

\section{Related Work}
In recent years, there has been increased interest at the intersection of NLP and the Legal domain, with work spanning  legal judgment prediction \cite{nchalkidis-etal-2019-neural,semo-etal-2022-classactionprediction,medvedeva2023legal} to Information Extraction\cite{holzenberger2023connecting,bommarito2021lexnlp} to Document analysis \cite{song2022multi,mamakas2022processing} to Text Generation {agarwal-etal-2022-extractive}. 

In the Information Extraction field, specifically in Named Entity Recognition (NER), \cite{amaral2023nlp} have focused on evaluating data agreements for compliance with European privacy laws using NLP techniques. In another study, \cite{smadu-etal-2022-legal} employed multi-task domain adaptation for NER within the legal domain, showing modest improvements in recall across Romanian and German languages. The work by \cite{barale2023language} asked language models to detect legal entity types. 
Additionally, NER has seen increased usage in the legal domain, including efforts to extract entities from court judgment documents in various jurisdictions \cite{kalamkar-etal-2022-named}. Additionally, \cite{au2022ner} presented E-NER; an Annotated Named Entity Recognition Corpus of Legal Text. However, these entity types, even in legal domain NER tasks, aren't specifically tailored for detecting legal violations and lack the complexity needed for this challenging task. Despite these advancements, existing research typically focuses on a standard set of entity types such as \textit{plaintiff} and \textit{defendant}, with limited exploration of more diverse or nuanced entities relevant to legal violations. Furthermore, these studies are limited in scope, often focusing on specific legal domains or industries.

NLI in the legal domain has gained significant attention in recent years. \cite{koreeda2021contractnli} explored NLI at the document level for contracts, while \cite{bruno2022lawngnli} introduced LawngNLI from US legal opinions. \cite{mathur2022docinfer} presented CaseHoldNLI and a document-level NLI model using optimal evidence selection. \cite{kwak2022validity} introduced a legal NLI dataset for the validity assessment of legal will statements and  \cite{kwak2023transferring} evaluated the validity of legal will statements across states, using three inputs—statement, condition, and law—to classify the relationship as \textit{support}, \textit{refute}, or \textit{unrelated}. Despite the increased interest, \cite{bernsohn2024legallens} is the first to introduce legal violation detection as a general NLI task across multiple domains.

Prior work has focused on domain-specific use cases, such as privacy protection \cite{amaral2023nlp, silvausingNlp2020, nyffenegger2023anonymity}, but these models lack the versatility needed to address the broad spectrum of legal violations across different contexts. \lens was the first to establish a cross-domain approach for detecting legal violations.

\section{Conclusion and Future work}

The LegalLens Shared Task demonstrated the potential of leveraging NLP techniques to address the challenge of legal violation detection across diverse domains. Despite the task's rapid timeline—less than two months from launch to completion—the significant participation of 87 individuals, organized into 38 teams, and the promising results underscore the community's interest and the relevance of this problem.

We call on the broader research community, particularly those in interdisciplinary fields, to contribute resources, methodologies, and diverse perspectives. Collecting and consolidating these perspectives will deepen our understanding of the complexities within this field. As we refine and build upon the LegalLens framework, we encourage diverse perspectives and innovative approaches that can address the challenges of this important task. Collaboration across disciplines will be crucial in advancing the state of the art in this important area.

The top models achieved a 0.416 F1 score in \ner (microsoft/deberta-v3-base) and 0.853 F1 score in \nli (phi3). However, a significant drop was observed in identifying the "Violated By" and "Violated On" entities, indicating room for improvement. This gap suggests the potential for integrating other information extraction techniques, even possibly from outside the legal domain.

Key questions remain unresolved: How will the techniques scale with larger language models and adapt to less-resourced languages? Can we enhance the granularity of legal entity interactions, particularly in more implicit scenarios? Additionally, how will these approaches generalize across broader legal domains and real-world applications?

\section*{Limitations}
 A challenge of identifying cases of legal violation in the open web is information sparsity. In other words, these cases do not present themselves in entirety, and in one place. Often times, the salient details of a case are spread across multiple sources on the web, and individually do not offer much insight into the case. It is only when these individual details are stitched together, do they afford themselves to a holistic understanding of the full story, and subsequent evaluation of the case.

\section*{Ethics Statement}
We strive to adhere to the \href{https://www.aclweb.org/portal/content/acl-code-ethics}{ACL Code of Ethics}. \\  \\ Bias and fairness in machine learning have been subjects of long-standing research. As we aim to develop more complex and impactful solutions to address the evolving media and world knowledge, we understand that this goes beyond merely developing or implementing ML algorithms. Inherent biases arise from datasets, task definitions, culture, and even researchers' beliefs and motivations. Addressing these biases effectively requires collaboration across disciplines. Our technology is designed to supplement, not replace, legal professionals, with responsible application and awareness of potential limitations and biases in automated systems. All data used in this research have been anonymized and stripped of personally identifiable information in compliance with relevant data protection regulations. The data utilized in this study are sourced from publicly available online platforms and do not infringe on any proprietary rights of individuals or entities.

\section*{Acknowledgements}
We would like to extend our gratitude to Darrow.ai for providing the dataset, computational resources, and domain expertise that made this research possible. Our thanks also go to the NLLP workshop for facilitating and helping to organize this shared task.

\bibliography{anthology,custom}

\appendix

\end{document}